\documentclass[10pt,twocolumn,letterpaper]{article}

\usepackage{cvm}
\usepackage{times}
\usepackage{epsfig}
\usepackage{graphicx}
\usepackage{amsmath}
\usepackage{amssymb}
\usepackage{multirow}

\usepackage[pagebackref=true,breaklinks=true,letterpaper=true,colorlinks,bookmarks=false]{hyperref}

\usepackage[table,xcdraw]{xcolor}
\usepackage[normalem]{ulem}
\usepackage{algorithm}
\usepackage{algorithmic}
\usepackage{subfigure}
\usepackage{caption}
\usepackage{mathrsfs}
\usepackage{comment}
\usepackage{hyperref}
\usepackage{authblk}

\newcommand{\keywords}[1]{{\bf \emph{Keywords: #1}}}

\cvmfinalcopy
\def\cvmPaperID{****} % *** Enter the cvm Paper ID here

\ifcvmfinal\fi
\begin{document}

%%%%%%%%% TITLE
\title{Pose-Controllable 3D Facial Animation Synthesis using Hierarchical Audio-Vertices Attention}

% This is how authors are specified in the conference style

% Author and Affiliation (single author).
%\author{Roy G. Biv\thanks{e-mail: roy.g.biv@aol.com}}
%\affiliation{\scriptsize Allied Widgets Research}

% Author and Affiliation (multiple authors with single affiliations).

\author[1]{Bin Liu\thanks{e-mail:nyliubin@nchu.edu.cn}} %
\author[1]{Xiaolin Wei}%\thanks{e-mail:ed.grimley@aol.com} %
\author[1]{Bo Li\thanks{e-mail: libonchu@outlook.com}}
\author[2]{Junjie Cao}
\author[3]{Yu-Kun Lai}

\affil[1]{Nanchang Hangkong University}
\affil[2]{Dalian University of Technology}
\affil[2]{Cardiff University}
% \affiliation{\scriptsize School of Mathematics and Information Science \\ Nanchang Hangkong University, Nanchang, China}

% Author and Affiliation (multiple authors with multiple affiliations)
% \author{Bo Li\thanks{e-mail: libonchu@outlook.com}\\ %
%         \scriptsize Starbucks Research %
% \and Ed Grimley\thanks{e-mail: ed.grimley@aol.com}\\ %
%      \scriptsize Grimley Widgets, Inc. %
% \and Martha Stewart\thanks{e-mail: martha.stewart@marthastewart.com}\\ %
%      \parbox{1.4in}{\scriptsize \centering Martha Stewart Enterprises \\ Microsoft Research}}

\maketitle
% \thispagestyle{empty}

%%%%%%%%% ABSTRACT
\begin{abstract}
    Most of the existing audio-driven 3D facial animation methods suffered from the lack of detailed facial expression and head pose, resulting in unsatisfactory experience of human-robot interaction. In this paper, a novel pose-controllable 3D facial animation synthesis method is proposed by utilizing hierarchical audio-vertex attention. To synthesize real and detailed expression, a hierarchical decomposition strategy is proposed to encode the audio signal into both a global latent feature and a local vertex-wise control feature. Then the local and global audio features combined with vertex spatial features are used to predict the final consistent facial animation via a graph convolutional neural network by fusing the intrinsic spatial topology structure of the face model and the corresponding semantic feature of the audio. To accomplish pose-controllable animation, we introduce a novel pose attribute augmentation method by utilizing the 2D talking face technique. Experimental results indicate that the proposed method can produce more realistic facial expressions and head posture movements. Qualitative and quantitative experiments show that the proposed method achieves competitive performance against state-of-the-art methods.
\end{abstract}


%%%%%%%%% BODY TEXT
\section{Introduction}

\begin{figure}
\centering
\includegraphics[width=1\linewidth]{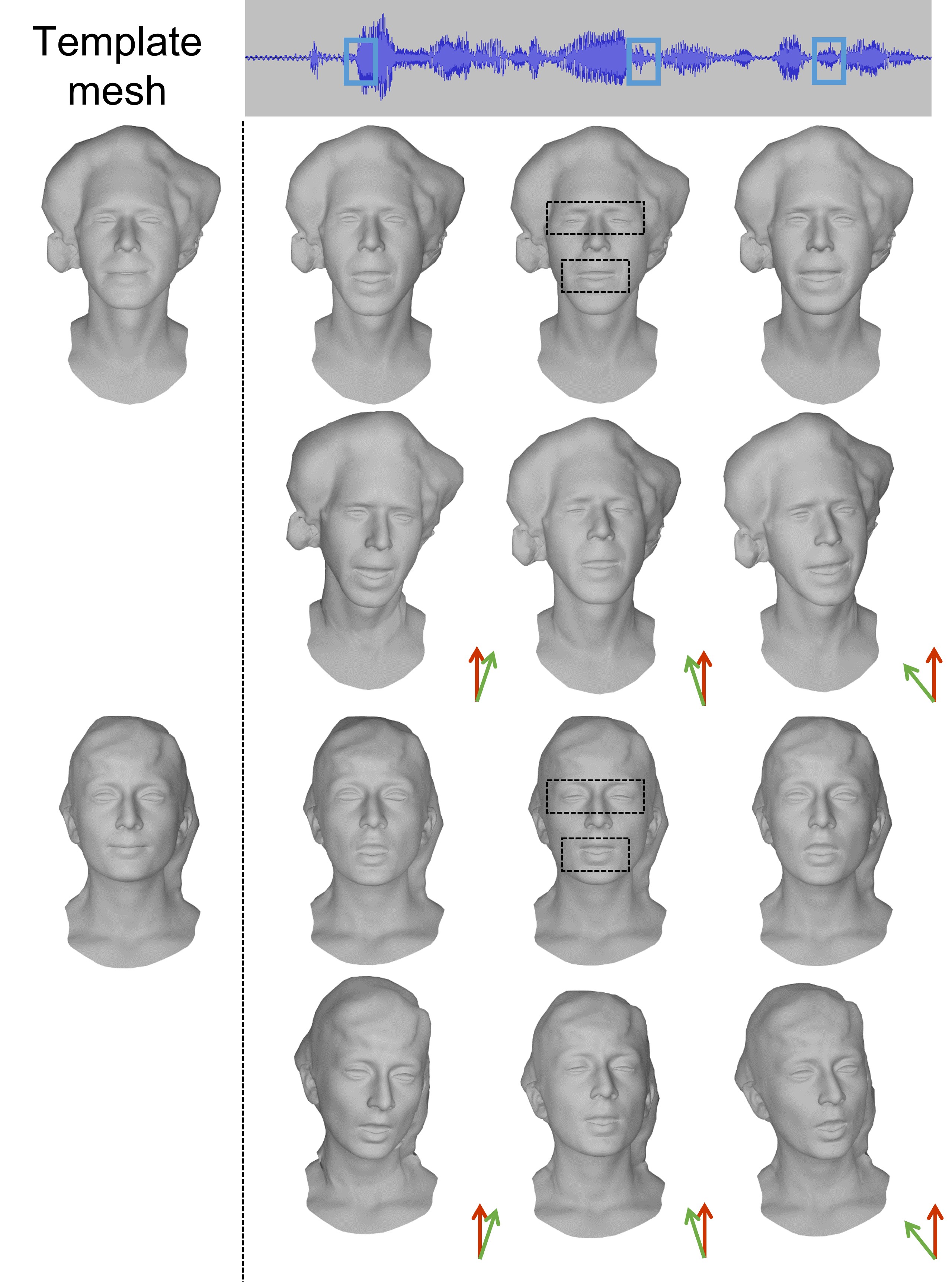}
\caption{Results of speech-driven facial animations generated by our method. Given the corresponding audio clips, our method generates reasonable mouth movement and the changes are smooth on the timeline. The red arrows indicate the orientation of the original template mesh, and the green arrows indicate the orientation of the current simulated mesh. The direction of each arrow is the vertex normal at the top of the nose. This rule applies to subsequent images in this paper.}
\label{ourresult}
\end{figure}

3D virtual digital human is a cutting-edge human-robot interaction scenario that combines artificial intelligence, natural language processing, computer vision and computer graphics. It can provide efficient and intelligent solutions for modern industries such as e-commerce, news broadcasting, film production and education. From the viewpoint of user experience design, realistic audio-driven 3D facial animation is one of the key techniques in 3D digital human research. Compared with 2D facial animation, 3D audio-driven animation can be more natural and vraisemblance.

% \R{In order to improve the user experience of human-computer interaction, some advances have been made in face alignment\cite{yang2016cascaded}, expression recognition\cite{wu2019weight} and texture synthesis\cite{kang2021competitive}.} However, the vraisemblance and naturalness of facial expression animation have not yet been achieved. Audio-driven 3D facial animation, therefore, is gaining popularity due to its ability to achieve the aforementioned objectives.

In general, audio-driven 3D facial animation methods can be divided into parametric methods~\cite{pham2018end,hussen2020modality} and non-parametric methods~\cite{cudeiro2019capture,liu2021geometry} according to the representation of 3D facial models. 

The advantage of the parametric approach is the convenience, however, they suffered from the following issues. They are not flexible enough and their accuracy is influenced by the linear assumption of parametric models. These aspects will result in the loss of facial expression details, which is important in the research of audio-driven 3D facial animation.
Benefiting from the end-to-end learning strategy, the non-parametric audio-driven 3D facial animation methods~\cite{cudeiro2019capture,liu2021geometry} have gained better flexibility and wider application. The key idea of these approaches is to learn a mapping function between audio representation space and 3D facial representation space through deep neural networks. However, most of the existing  audio-driven 3D facial animation methods suffered from the lack of detailed facial expression and head pose, and resulted in inconsistent synthesis results to actual human face animation. 
% Therefore, the intermediate features extracted from the speech and mesh information will play a significant role during the network training process. 
On the one hand, most of the related methods~\cite{cudeiro2019capture,liu2021geometry,richard2021meshtalk} utilize encoder-decoder structured networks to build the mapping between audio and facial spaces. However, the encoder networks in the above methods only pay attention to the global features of audio or face mesh, which result in facial animation lack of detailed expression. 
% As a result, they ignore details usually controlled by local features.
On the other hand, due to that the popular dataset, such as VOCASET~\cite{cudeiro2019capture} and MeshTalk\cite{richard2021meshtalk}, do not have 
% the attribute of head pose 
a range of motion sequence for head pose corresponding to audio, most of the existing methods can only produce facial animation without pose variance, and cannot achieve pose-controllable animation results.

% Decoder networks aim at learning the vertex offsets relative to the template mesh, without taking into account the changes of head pose. These two reasons would cause a lack of detail and movement stiffness during facial expression simulation.

To solve the above issues, 
%as well as to increase realism, 
a novel pose-controllable 3D facial animation synthesis method based on hierarchical audio-vertex attention is proposed in this paper. 

First, in order to synthesize facial animation with more details, we propose to extract hierarchical audio-vertex features rather than merely global features used in most existing works. Two networks are designed to extract the global and local features of the audio signal respectively. Global feature implies the overall facial expression, while local feature learns the latent movement feature of each vertex with respect to the global facial expression. Then, a novel audio-vertex hierarchical feature is designed by combining both local and global audio features with the Fourier embedding features of the index of each vertex of the facial mesh. Next, to take full advantage of the topology of face model, we utilize a graph convolutional neural network to further fuse the intrinsic geometry features. Finally, the displacement of each vertex corresponding to the input audio will be predicted.

Second, in order to synthesize realistic facial animation with pose variance, we propose to establish a novel pose attribute augmentation method based on 2D talking face technique, and then propose an adaptive head pose prediction network to produce the realistic head movement corresponding to the input audio.
Benefiting from huge amount of training data with diverse head postures in 2D videos, the synthesis of 2D talking face animation has achieved great advantages in the past decades. Although 2D face animation cannot render realistic facial images as well as 3D synthesis methods especially in cases with obvious occlusions, 2D methods can produce diverse head postures.
%benefiting from huge amount of training data with various head postures in 2D videos.
Inspired by the above analysis, a novel pose attribute augmentation method is proposed by predicting the 3D posture of the synthesized 2D face animation with the same audio input.

% generate a natural variation of head movement by adding pose attributes to the original audio-driven 3D facial dataset~\cite{cudeiro2019capture} (VOCASET). 
%\R{In addition, we use a new network that can predict the head pose to generate a natural variation of head posture. To train the network, we add pose attributes to an audio-driven 3D facial dataset~\cite{cudeiro2019capture} (VOCASET), using a method proposed by ~\cite{zhou2020makelttalk} and ~\cite{feng2021learning}.} 
Extensive experiments demonstrate that our method ensures high-quality facial animation with detailed expression while adding more realistic head pose variations. In summary, the main contributions of the work include:

\begin{itemize}
\item[$\bullet$] A novel hierarchical audio-vertex feature is proposed by integrating the global facial expression and the local latent movement feature of each vertex. 
Furthermore, a graph convolutional neural network is utilized to take full advantage of the intrinsic geometric regularization.

\item[$\bullet$] 
A novel attribute augmentation method is proposed to add posture attribute to the popular datasets, including VOCASET\cite{cudeiro2019capture} and MeshTalk\cite{richard2021meshtalk}. An adaptive head pose prediction network is proposed to produce the realistic head movement corresponding to the audio.
% We propose a novel process for generating 3D head poses and use it to add pose attributes to the VOCASET\cite{cudeiro2019capture}, thus building a more practical 3D audio-driven facial dataset. Based on this dataset, we realize the variation of head poses in the process of facial animation.

\item[$\bullet$] Qualitative and quantitative experiments on VOCASET and MeshTalk demonstrate that our approach outperforms the state-of-the-art methods. In addition, our method can also perform well on unseen subjects and cross-linguistic applications.
\end{itemize}

\section{Related Work}
Existing audio-driven 3D facial animation techniques can be classified into two categories in general, parametric methods and non-parametric methods. In this section, we review some work that is highly related to us.

\subsection{Parametric facial animation synthesis}
Early approaches~\cite{hong2002real,zhang2010facial} are mainly based on some facial animation systems such as IFACE\cite{hong2001iface} and FACS\cite{ekman1978facial}. However, those synthetic results are very coarse and lack of realism. To improve the representation accuracy of face representation, researchers have proposed a more mature 3D face deformable modeling method (3DMM)\cite{blanz1999morphable}. The main idea is decoupling facial representation space into a parametric model related to expression, identity, and head posture. This decomposition strategy has greatly improved the accuracy and realism of face representation and is widely used in the field of facial animation synthesis\cite{li2016expressive,pham2018end,kim2019multi,zhang20213d,wu2021voice2mesh}. The main differences among these methods are the speech encoder and coefficient regression model related to the parametric model. Although Zhang \etal~\cite{zhang20213d} used a generative adversarial network to produce a head pose sequence for a given audio, the performance of detailed animation is still restricted to the parametric representation. And it is difficult to achieve real-time. We can move this reference to the next paragraph.

To enhance the realism in facial animation, other researchers\cite{hussen2020modality,liu2015video,pham2017speech,huang2018visual,richard2021audio,zhang20213d} attempt to predict the semantic parameters of head posture from captured face video. For example, Hussen \etal~\cite{hussen2020modality} propose a joint audio and video driving 3D face animation synthesis system. They first use convolutional neural networks to extract visual features and audio features from video and audio respectively, then regress the head pose parameters from the visual features, while predicting the identity parameters and expression parameters of the face from the fused visual and audio features, and finally reconstruct the 3D face by 3DMM. 

Although parametric methods obtain good results, they still have some limitations. These methods mainly depend on the performance of 3DMM, however, the main technique of 3DMM is principal component analysis (PCA) which makes it difficult to represent the details of the face, such as wrinkles. Therefore, some researchers begin to use non-parametric face representation to realize face animation generation.

\subsection{Non-parametric facial animation synthesis}
Karras \etal~\cite{karras2017audio} proposed a non-parametric facial animation synthetic method for the professional actor. They first use a neural network to learn the phonological features from the actor's speech and then use a decoder to directly predict vertex positions from the speech feature. 

In order to model speaker-independent facial animation, some researchers\cite{cudeiro2019capture,liu2021geometry,chai2022speech,fan2022joint} try to improve the generalization capability of decoder. Daniel \etal~\cite{cudeiro2019capture} use PCA to initialize the face representation latent space, then utilize neural network to update it. Liu \etal~\cite{liu2021geometry} first employ nonlinear autoencoder network to learn the geometric prior of face representation latent space in a face mesh dataset, then the geometric prior is used to constrain the face representation space which is learned from speech domain. The idea of Chai \etal~\cite{chai2022speech} is to decode the deformation gradient of each vertex in the mesh with respect to the template mesh from the face representation space\cite{sumner2004deformation}. In addition, Fan \etal~\cite{fan2022joint} adopt not only audio features but also text features to generate facial animation. However, these methods only consider the global feature of audio, which may result in a smooth result.

To achieve more naturalistic animation effects, Lahiri \etal~\cite{lahiri2021lipsync3d} propose a learning framework for personalized 3D face animation synthesis based on the video. They use face detection, 3D alignment, and mesh deformation techniques to make the predicted face with similar poses and expressions in a reference video. Fan \etal~\cite{fan2022faceformer} use transformer to explore the temporal correlation on the speech and facial animation sequences. 
%\R{Richard \etal~\cite{richard2021meshtalk} utilize} a large amount of training data to allow the neural network to establish a categorical association between the speech space and the facial expression space. They further control the reconstruction loss of the upper and lower parts of the face respectively with the help of masks to achieve eye movement. They have achieved more realistic results compared to VOCA\cite{cudeiro2019capture} and have made great progress with details like eye motion. In addition, 
Richard \etal~\cite{richard2021meshtalk} use a two-stage facial animation method. They first learn a facial motion representation space from audio and corresponding face mesh. Then, an autoregressive neural network is used to infer the next facial motion state from the learned facial motion space.

However, most of the existing non-parametric methods ignored the local spatial attention of audio features and resulted in lack of detailed animation. In addition, as the existing datasets including VOCASET and MeshTalk do not have posture attribute, most of the existing methods cannot produce realistic talking face with fluent head movement.

%\B{However, we believe that much of their results are due to their prior facial mask, and that their training strategy and loss design make this mask indispensable. Compared with VOCA, their progress is to choose to use multiple features to represent the motion of a mesh, but both methods only stay at the mesh level. In addition, MeshTalk\cite{richard2021meshtalk}'s training process is split into two steps and it has a large model, which results in both training and testing taking a long time. }

\begin{figure*}
\centering
\includegraphics[width=1\linewidth]{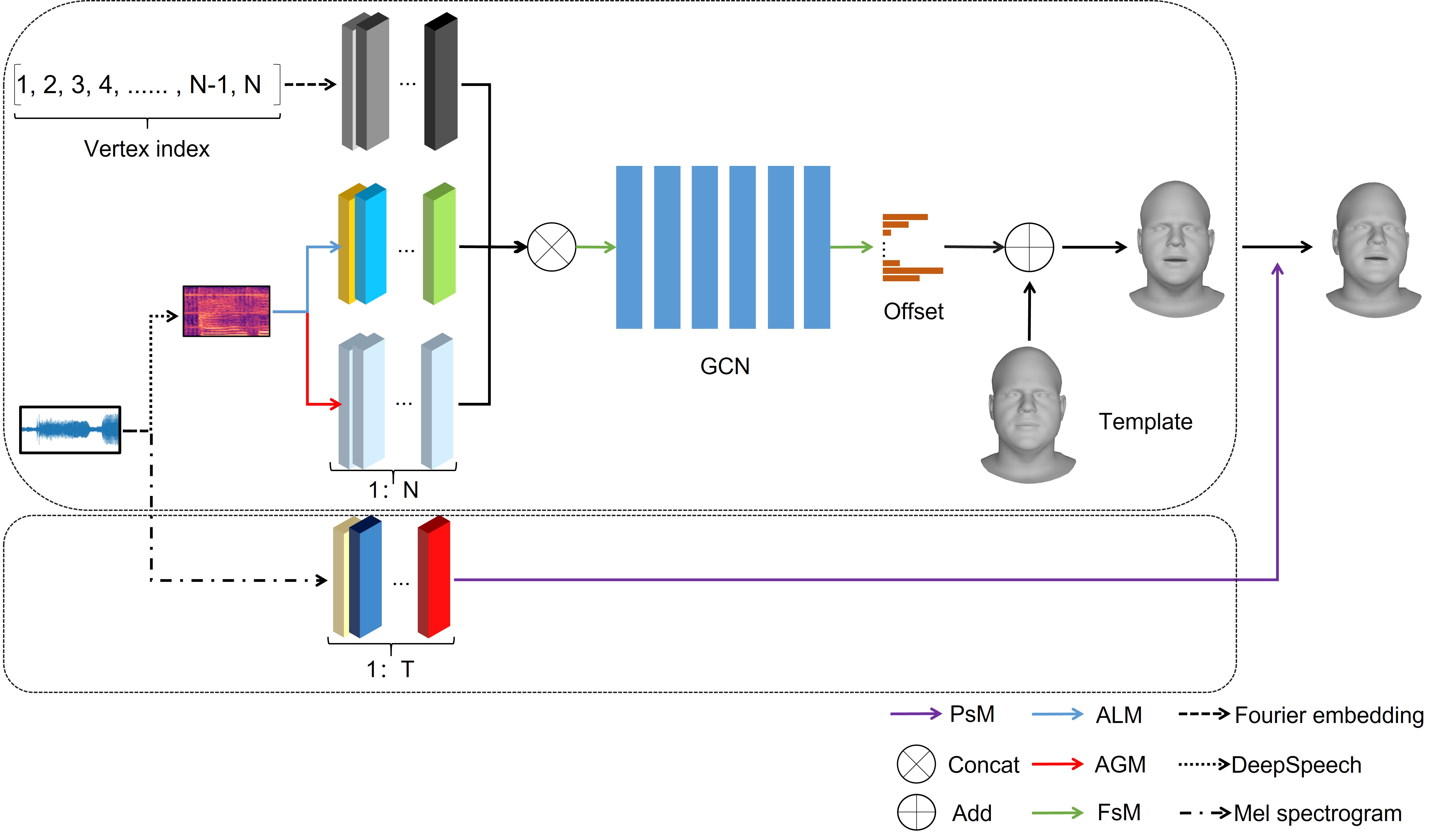}
\caption{First, the input audio is transformed into DeepSpeech\cite{hannun2014deep} features and Mel spectrogram. The former is sent to AGM and ALM to obtain local and global features of input audio for each vertex, and the latter is sent to the pose network to obtain the pose corresponding to the audio. Then the local and global features are concatenated with the Fourier embedding of the index of each vertex of the 3D template and fed to the GCN\cite{2018Weisfeiler} to obtain the offsets for each vertex. Adding the vertex displacement to the mesh template gives the talking 3D human head. Finally we process the speaker head with the pose for each frame obtained from the pose network to get a more realistic animation.}
\label{model}
\end{figure*}

\section{Technique Approach}
Given a template mesh and an input audio, our goal is to generate realistic and detailed 3D facial expression animation with fluent poses consistent to the input audio. To figure it out, the proposed method is composed of the following two stages: the first stage aims to predict the displacement of each vertex related to the given audio, and the second stage is designed to predict the head poses corresponding to the audio. The architecture of the proposed method is illustrated in Fig.~\ref{model}. The first stage contains four modules: Audio Local Module (ALM), Audio Global Module (AGM), Mesh Local Module (MLM) and Fusion Module (FsM), and the second stage consists of Pose Module (PsM). Specifically, ALM is used to extract local features of audio corresponding to each vertex of the template mesh. The global features of audio are extracted by AGM, which represents the emotional information. MLM is conducted by Fourier embedding, which is performed on the vertex index of the template mesh. FsM is a graph convolutional neural network, which is used to fuse the features outputted by ALM, AGM, and MLM. In the second stage, we use PsM to predict the variation of head posture during the process of facial animation synthesis. Each module will be presented in detail.
% Our goal is to generate real and detailed 3D facial expression animation with poses from audio. In order to figure it out, we design the following five modules: Audio Local Module (ALM), Audio Global Module (AGM), Mesh Local Module (MLM), Fusion Module (FsM) and Pose Module (PsM). The first four modules form the first stage of our proposed network architecture, which aim to predict details. As the second stage, the last module is used to predict face movement. The whole architecture of our proposed method is shown in Fig.~\ref{model}. ALM is used to extract local features of audio corresponding to each vertex of template mesh. The global features of audio are extract by AGM, which represents the emotional information. MLM is completed by Fourier embedding, which is performed on the vertex index of template mesh. FsM is consisted of graph convolutional neural network, which is used to fuse the features outputted by ALM, AGM and MLM. We use PsM to predict the variation of head posture during the process of facial animation synthesis. Next, each module will be presented in detail.

\subsection{Problem definition}
% In this paper, we organize data in the following form,  $\{(\mathbf{I},\mathbf{d}_i,\mathbf{y}_i,\mathbf{m}_i,\mathbf{p}_i)\}^T_{i=1}$. Here, index $i$ refers to a specific frame, and $T$ is the total number of frames. $\mathbf{d}_i\in\mathbb{R}^{W\times D}$ is the speech feature window centered at the $i$th frame generated by DeepSpeech~\cite{hannun2014deep}, where $D$ is the number of phonemes in the alphabet plus an extra one for a blank label and $W$ is the window size. $\mathbf{I}\in\mathbb{R}^{N\times 1}$ means the vertex index of the corresponding template mesh, and $N$ is the number of vertices of the mesh. $\mathbf{y}_i\in\mathbb{R}^{N\times 3}$ denotes the ground truth spatial coordinate for facial animation at each frame. $\mathbf{m}_i\in\mathbb{R}^{F\times L}$ represents the Mel spectrogram transformed from the raw waveform at the $i$th frame, where $F$ is mel filterbanks and $L$ is length. $\mathbf{p}_i\in\mathbb{R}^{N\times 3}$ denotes the ground truth for head poses at each frame. At last, let $\hat{\mathbf{y}_i}\in\mathbb{R}^{N\times 3}$ and $\hat{\mathbf{p}_i}\in\mathbb{R}^{N\times 3}$ denotes the output of our model for the input $\mathbf{d}_i,\mathbf{m}_i$ with template $\mathbf{I}$.

In this paper, we organize data in the following form,  $\{(\mathbf{I},\mathbf{y}_i,\mathbf{p}_i,\mathbf{d}_i,\mathbf{m}_i)\}^T_{i=1}$. Here, index $i$ refers to a specific frame, and $T$ is the total number of frames. $\mathbf{I}\in\mathbb{R}^{N\times 1}$ means the vertex index of the corresponding template mesh, and $N$ is the number of vertices of the mesh. $\mathbf{y}_i\in\mathbb{R}^{N\times 3}$ denotes the ground truth spatial coordinate for facial animation at each frame. $\mathbf{p}_i\in\mathbb{R}^{N\times 3}$ denotes the ground truth for head poses at each frame. $\mathbf{d}_i\in\mathbb{R}^{W\times D}$ is the speech feature window centered at the $i$th frame generated by DeepSpeech~\cite{hannun2014deep}, where $D$ is the number of phonemes in the alphabet plus an extra one for a blank label and $W$ is the window size. $\mathbf{m}_i\in\mathbb{R}^{F\times L}$ represents the Mel spectrogram transformed from the raw waveform at the $i$th frame, where $F$ is mel filterbanks and $L$ is length. At last, let $\hat{\mathbf{y}_i}\in\mathbb{R}^{N\times 3}$ and $\hat{\mathbf{p}_i}\in\mathbb{R}^{N\times 3}$ denotes the output of our model for the input $\mathbf{d}_i,\mathbf{m}_i$ with template $\mathbf{I}$.

\subsection{Hierarchical features with movement details}
To enable our method to produce fine facial motion, we construct an architecture that produces a hierarchical feature for each vertex on the template mesh. It is composed of the ALM, AGM, and MLM, which are closely related to the local audio features, the global audio features, and the local mesh features. To further optimize the hierarchical feature, we utilize the FsM which encodes the topology information of the template mesh. Afterward, these hierarchical features are fed to a multilayer perceptron whose output is the displacement of each vertex in the template mesh.

\textbf{Audio Local Module (ALM).} The role of the ALM is to generate local speech features corresponding to each vertex, with the expectation of generating refined facial movements. It contains 5 convolutional layers and 4 multilayer perceptrons. Convolutional kernel with size 4 and step 1 is applied to learn variations of local features in the time dimension. To obtain features that are closely related to texts, we use DeepSpeech features $\mathbf{d}_i$ as the input of ALM, which predicts the probability distribution of characters. Therefore, ALM can perceive the local movement of facial animation. Detailed structure is shown in the supplementary material.

\textbf{Audio Global Module (AGM).} AGM is used to extract the audio global feature, which may represent the speaking style or habit, driving the motion of vertices in template mesh at a high level. AGM contains 4 convolutional layers and 2 multilayer perceptrons. Similar to ALM, the convolutional operation is performed in the time dimension, with kernel size 3 and step 2. The input of AGM is also $\mathbf{d}_i$, and supplementary material displays the detailed operation process.

% \R{The fully connected layers are not up-dimensioned here. After passing through the fully connected layers, the resulting features are copied in $N$ copies. Here $d$ is still the input to the global net, so the flow of the global net can be expressed as:

% \begin{equation}
% \begin{aligned}
% x_{global} = Copy(MLP_{G}(E_{G}(d)))
% \label{F3}
% \end{aligned}
% \end{equation}

% where $MLP_{G}(E_{G}(d))\in\mathbb{R}^{S\times 1}$, $Copy()$ is a copy function, and $x_{global}\in\mathbb{R}^{S\times N}$. Eventually, we swap the second and third dimensions of $x_{global}$ in order to keep $x_{local}$ and $x_{global}$ in the same shape.}

\textbf{Mesh Local Module (MLM).} We use MLM to extract the local feature of the vertex in the template mesh. In practice, MLM represents fast Fourier embedding. As the topology structure of all meshes in the VOCASET is consistent, vertex index $\mathbf{I}$ is mapped into a high dimensional space by Fourier transform in this paper. Notably, $\mathbf{I}$ is first normalised to [-1, 1]. The reason that we do not use 3D coordinates as the input of Fourier embedding is that the index is permanent and has a very clear semantic attribute. For example, it represents a point on a nose or lips. In addition, this invariance contributes to the robustness and generalization of our method.

% \R{Once we have the indexes, we still need to embed them and put them into the network. Here we have chosen the most common Fourier embedding, where each index is normalised to [-1, 1] and then embedded in 40 dimensions. This feature concatenates the index itself is 41 dimensions, we represent each index by the vector $I\in\mathbb{R}^{N\times 41}$.}

\textbf{Fusion Module (FsM).}
Once hierarchical features are obtained by the ALM, AGM, and MLM modules, they are fed into FsM module by concatenation operation. FsM contains 8 graph convolutional layers~\cite{2018Weisfeiler}, which makes full use of the topology structure of template mesh, especially the adjacency relation among vertices. To better propagate hierarchical features, we use residual learning in each layer, followed by a LeakyReLU activation function. After the hierarchical features are optimized by the graph convolutional network, they are passed to a multilayer perceptron that outputs the displacement of each vertex, without any activation function. The elaborated network setting is shown in the supplementary material.

% \R{
% For these topologically consistent models, graph convolution works best. Our graph convolution network has a total of eight layers, each of which is a single layer of graph convolution. We use the GraphConv function provided by pytorch3d, which takes the following concrete form:

% \begin{equation}
% \begin{aligned}
% X_i^n = LeakyReLU(MLP_0^{n-1}(X_{i}^{n-1})+gather(MLP_1^{n-1}(X_{i}^{n-1})))
% \label{F5}
% \end{aligned}
% \end{equation}

% where $X_i^n$ is the $ith$ vertex output of the $nth$ GCN layer. Specifically, $X^0$ is the input to the first layer. $MLP_0^{n-1}$ and $MLP_1^{n-1}$ are two fully connected layers respectively, and the $gather()$ function is an aggregate function that adds the features of a vertex's neighboring vertices to its own features according to the adjacency of the template mesh. And the input to the layer one is:

% \begin{equation}
% \begin{aligned}
% X^0 = cat(I, x_{local}, x_{global})
% \label{F6}
% \end{aligned}
% \end{equation}

% where $cat()$ is a concatenation function. After passing through the final layer of GCN, there is another fully connected layer mapping the features to the displacement of each vertex, without any activation function.}

\subsection{High reality with pose variation}\label{sec:attri}
In this section, we first introduce how to add a pose attribution to the original audio-driven 3D facial movement dataset VOCASET~\cite{cudeiro2019capture}. Then, PsM is proposed to predict head posture variation by supervised learning.

\textbf{Attribution Augmentation.}
VOCASET~\cite{cudeiro2019capture} contains a collection of audio-4D scan pairs captured from 6 female and 6 male subjects, each subject speaks 40 sentences, 3D facial movements are captured at a frame rate of 60FPS and are registered well using the publicly available generic FLAME model~\cite{li2017learning}.  All meshes are in a ``zero pose'' state. In addition, it also contains a figure for each subject. However, this dataset has no variation in head posture. 

To generate more meaningful facial posture movement, a novel pose attribute augmentation method is proposed in this section. The basic idea of the proposed method is that 2D
talking face animation methods can produce more realistic postures of head images benefiting from huge amount of training data, then we can predict the 3D pose parameter from the 2D video with the same audio input as the pose attribute of 3D face models in VOCASET.
The proposed method is composed of two strategies. First, given the figure of one subject in VOCASET and its corresponding audio, we use the method described in~\cite{zhou2020makelttalk} to synthesize a facial video with pose variation. 
% The reason that we choose the above method is it can disentangle the audio signal into speech content and speaker information and focus more on the behavior of each subject. We propose using a 2D generation method for poses rather than 3D, as this would provide a more realistic and reliable result. There are many videos of speech available with variations in the head posture that can be used for training. 
Next, we utilize the method proposed by~\cite{feng2021learning} to predict the pose parameters of the head, which is consistent with the FLAME model~\cite{li2017learning}. To obtain more continuous poses, a Gaussian filter with standard deviation $1$ and window size $29$ is used to smooth the estimated pose parameters along the time sequence.

\textbf{Pose Module (PsM).}
In this section, we design a pose module to predict the head posture corresponding to an audio clip. Mel spectrogram is a spectrogram of frequencies converted to the Mel scale, which is more intuitive to humans and can help us to get the poses from the audio. LSTM is very effective in dealing with the time series and can extract some useful pose information about human habits from a given audio clip. Therefore, we first extract the Mel spectrogram feature of the audio clip. Then the Mel spectrogram $\mathbf{m}_{1:t}$ is fed into a deep neural network, which contains 7 convolutional layers, 2 LSTM layers, and 1 multilayer perceptron. Next, the pose displacement which is represented by a 3D rotation vector between the first frame ("zero pose") and the current frame of the audio clip is output by a deep neural network. 
% Detailed parameter settings are shown in the supplementary material.

\begin{figure*}
\centering
\includegraphics[width = 1\linewidth]{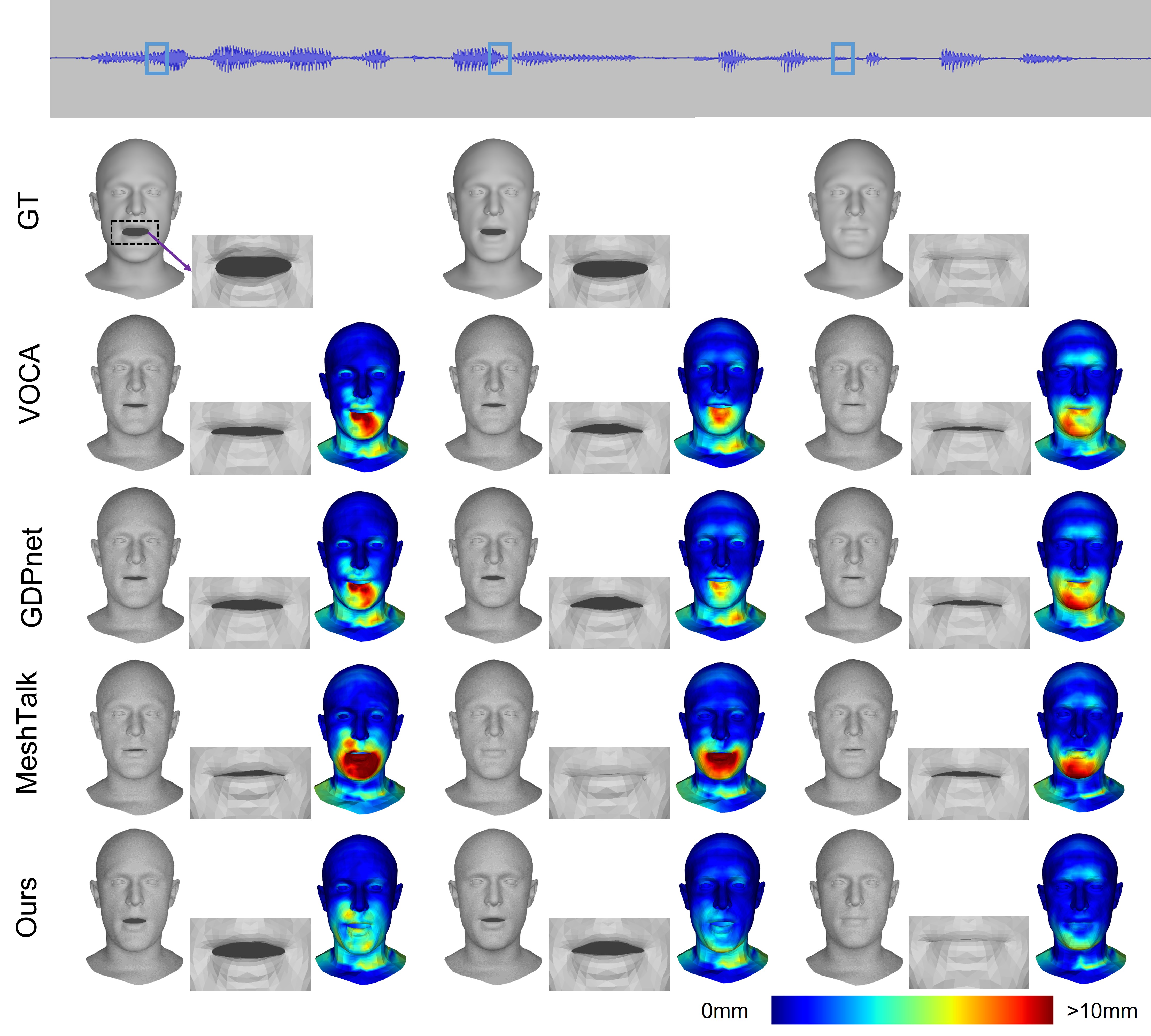}
\caption{Experimental results on noiseless audio in VOCASET\cite{cudeiro2019capture}. We enlarged the mouth area, and the color maps give the distribution of vertex-to-vertex distance errors (unit: millimeter). We can find that our method has better ability in preserving details, such as the lips and chin.}
\label{numevaerrormap1}
\end{figure*}

\begin{figure*}
\centering
\includegraphics[width=1\linewidth]{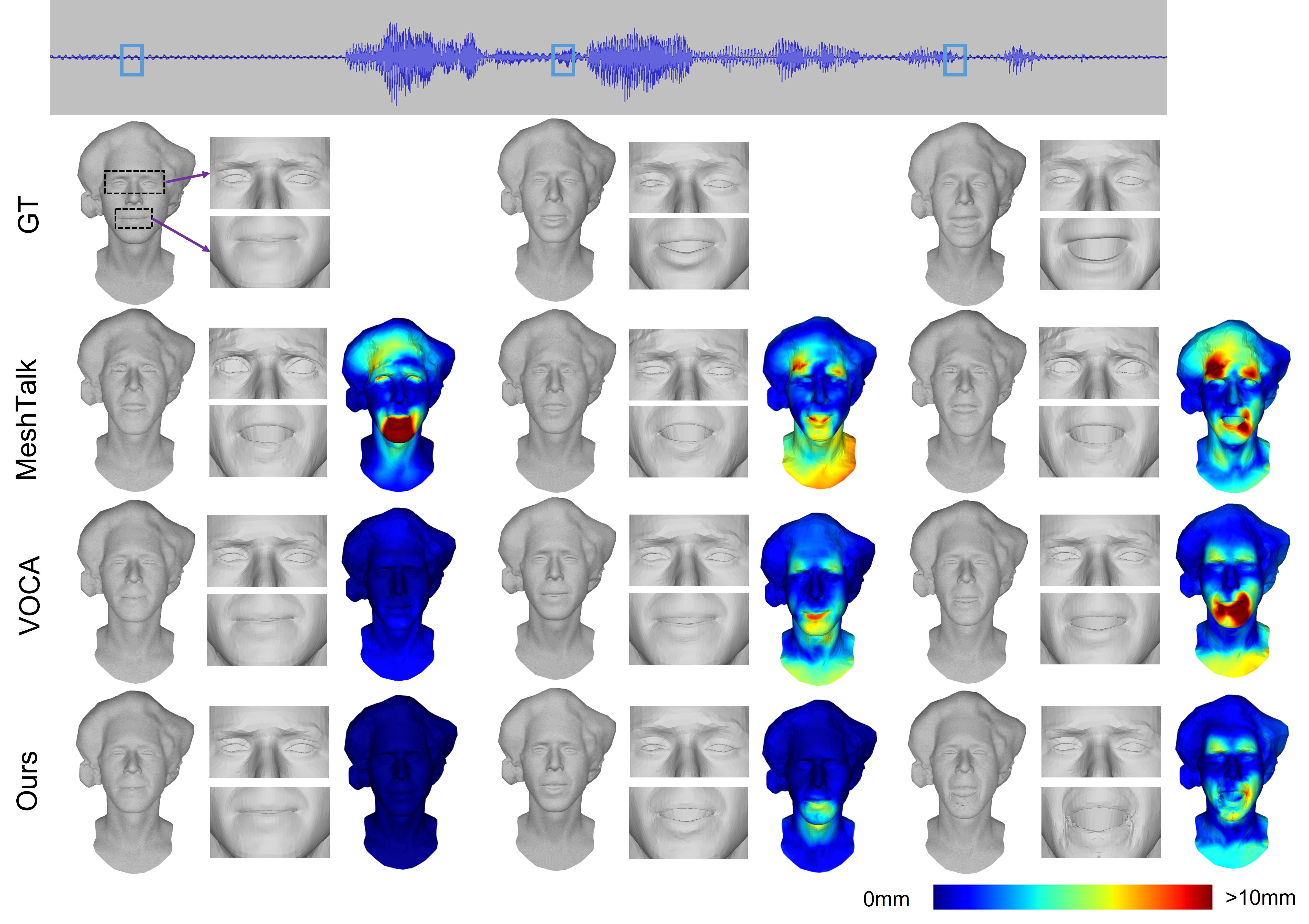} 
\caption{Experimental results on noiseless audio in MeshTalkSet\cite{richard2021meshtalk}. We enlarged the mouth and eye area, and the color maps give the distribution of vertex-to-vertex distance errors (unit: millimeter). We can find that our method has better ability in preserving details, such as the lips and eyes.}
\label{numevaerrormap2}
\end{figure*}

\subsection{Loss function}
We now discuss the loss function used in training. The loss $\mathscr{L}_1$ in the first stage consists of two parts, reconstruction loss $\mathscr{L}_r$ and velocity loss $\mathscr{L}_v$ respectively:
\begin{equation}
\mathscr{L}_1 = \mathscr{L}_r + \lambda \mathscr{L}_v,
\label{F8}
\end{equation}
where $\lambda$ is the weight parameter to balance the above two items. In experiments we set its value to 10.

The error metric for reconstruction loss $\mathscr{L}_r$ between the ground truth $\mathbf{y}_i$ and the predicted 3D model $\hat{\mathbf{y}}_i$ can be defined as the sum of absolute vertex-to-vertex distances:
\begin{equation}
\mathscr{L}_r = \Vert\mathbf{y}_i - \hat{\mathbf{y}}_i\Vert_1,
\label{F9}
\end{equation}
where $||\cdot||_1$ represents the sum of absolute values of each element. The velocity loss is used to induce temporal stability, which considers the smoothness of prediction and ground truth in the sequence context. It can be written as
\begin{equation}
\mathscr{L}_v = \Vert(\mathbf{y}_i - \mathbf{y}_{i-1}) - (\hat{\mathbf{y}}_{i} - \hat{\mathbf{y}}_{i-1})\Vert_1.
\label{F10}
\end{equation}

The loss function $\mathscr{L}_2$ used in the second stage is the least square loss between the predict posture vector and the ground truth obtained in Sec.~\ref{sec:attri}, which is 
\begin{equation}
\mathscr{L}_2 = \Vert\mathbf{p}_i-\hat{\mathbf{p}}_i\Vert_2^2 \, . \\
\label{F11}
\end{equation}
%where, $y_i$ is ground truth of the the mesh vertex motion at frame $i$, and $f_i$ is predicted values for the $ith$ frame, $y_i, f_i \in \mathbb{R}^{N\times 3}$. $\lambda$ is is the weighting of velocity loss. 

% \begin{figure*}
% \centering
% \includegraphics[width = 1\linewidth]{data/compare/compare1.jpg}
% \caption{Comparisons on another example for clean audio as input.}
% \label{numevaerrormap2}
% \end{figure*}

\begin{figure*}
\centering
\includegraphics[width = 1\linewidth]{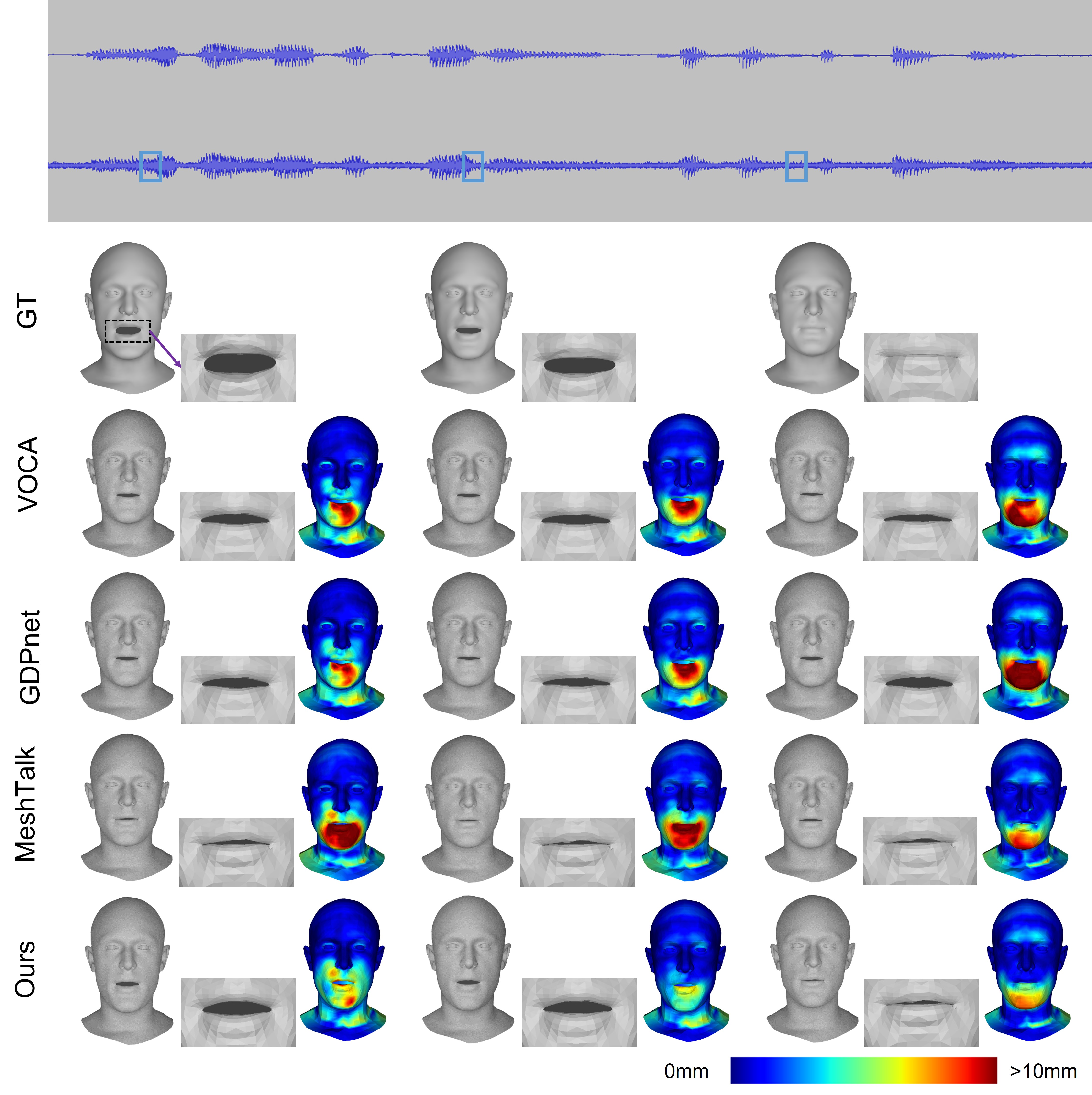}
\caption{Results on audio with Gaussian noise in VOCASET\cite{cudeiro2019capture}. We enlarged the mouth area, and the color maps give the distribution of vertex-to-vertex distance errors (unit: millimeter). It can be seen that our method still has good performance.}
\label{noiseerrormap}
\end{figure*}

\begin{figure*}
\centering
\includegraphics[width = 1\linewidth]{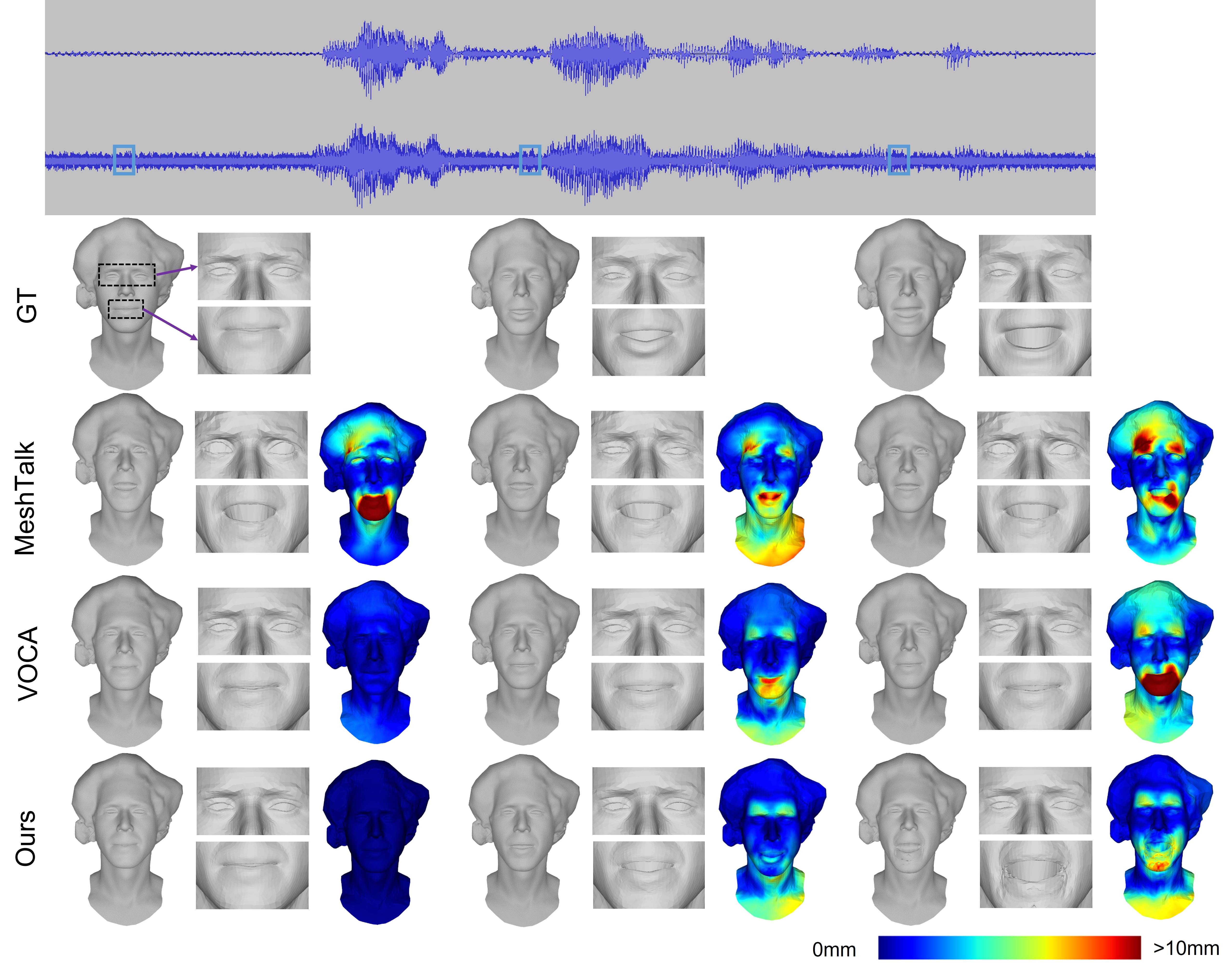}
\caption{Results on audio with Gaussian noise in MeshTalkSet\cite{richard2021meshtalk}. We enlarged the mouth and eye area, and the color maps give the distribution of vertex-to-vertex distance errors (unit: millimeter). It can be seen that our method still has good performance.}
\label{noiseerrormap2}
\end{figure*}

\begin{figure*}
\centering
\includegraphics[width=1\linewidth]{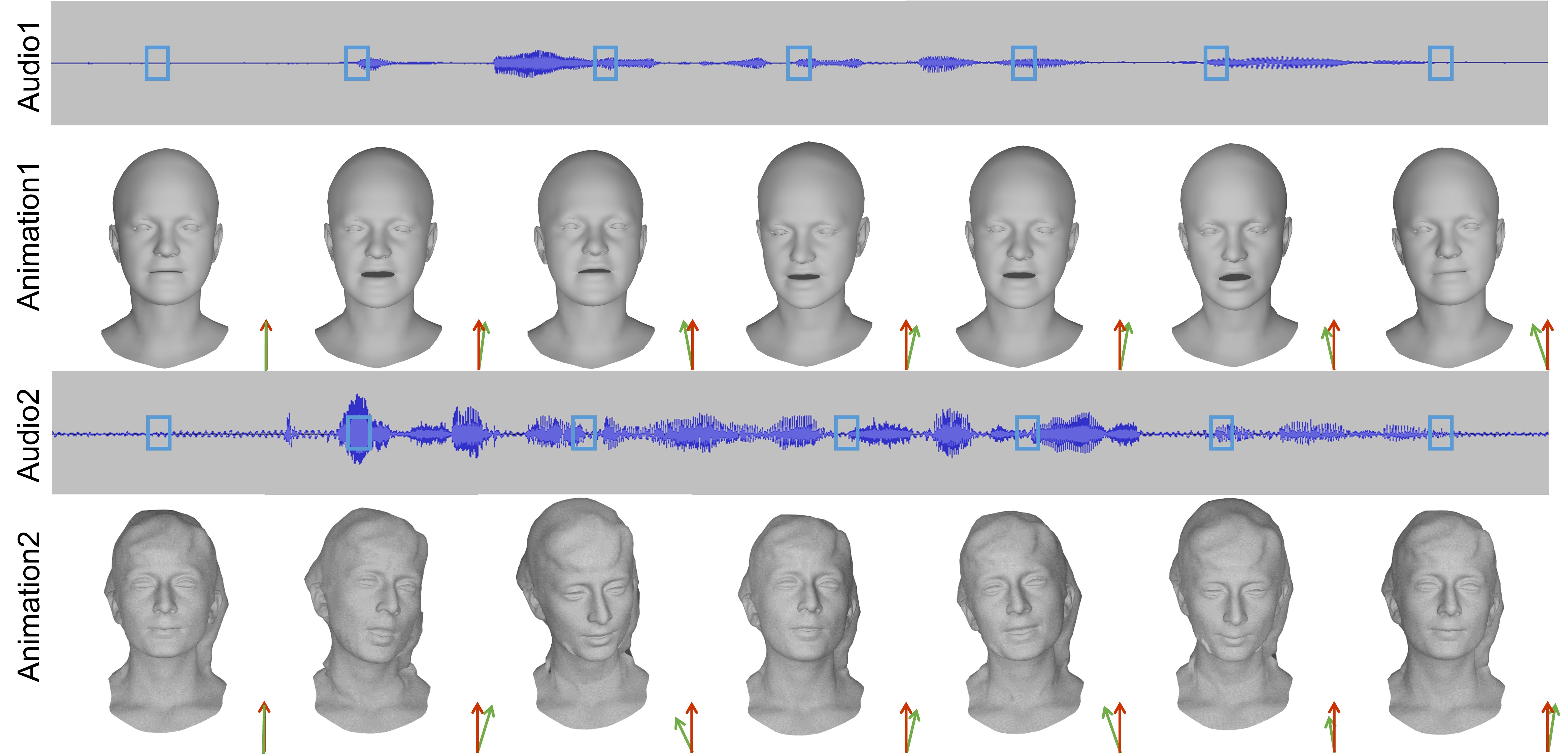} 
\caption{Examples of head posture changes.}
\label{poseeva}
\end{figure*}

\section{Experiment}
In this section, we first introduce the implementation details of our method. Then  qualitative and quantitative experiments are conducted to demonstrate the effectiveness of our method. 
%We conduct some qualitative and quantitative experiments in the following sections. 
\subsection{Implementation details}
We train the network by using the Adam optimization algorithm running on a PC equipped with an Intel(R) Core(Tm) i9-9900K CPU @ 3.2 GHz, 32GB RAM, and a GeForce RTX 3090 GPU. The epochs and batch size are set to 50 and 64, respectively. In the second stage, the epochs and batch size are set as 1 and 8 and the time step is set to 30 in the LSTM neural network. The learning rate remains the same at the two stages and both are 1e-4.

\begin{table}[t]
\centering
\caption{Comparison with other methods in different datasets on noiseless audio. $E_{vl}$ and $E_{ve}$ represent the errors in lip and eye regions, respectively.The best results are highlighted (unit: millimeter). }
\begin{tabular}{c|cc|cc}
\hline
\multirow{2}{*}{}&\multicolumn{2}{c|}{VOCASET\cite{cudeiro2019capture}}&\multicolumn{2}{c}{MeshTalkSet\cite{richard2021meshtalk}} \\ \cline{2-5} 
&\multicolumn{1}{c|}{$E_{vl}$}&$E_{ve}$&\multicolumn{1}{c|}{$E_{vl}$}&$E_{ve}$\\ \hline
VOCA\cite{cudeiro2019capture}&\multicolumn{1}{c|}{6.384}&1.993&\multicolumn{1}{c|}{5.075}&3.224\\ 
GDPnet\cite{liu2021geometry}&\multicolumn{1}{c|}{6.062}&1.986&\multicolumn{1}{c|}{-}&-\\ 
MeshTalk\cite{richard2021meshtalk}&\multicolumn{1}{c|}{6.449}&2.097&\multicolumn{1}{c|}{6.385}&4.462\\ 
Ours&\multicolumn{1}{c|}{\textbf{5.261}}&\textbf{1.814}&\multicolumn{1}{c|}{\textbf{4.676}}&\textbf{3.034}\\ \hline
\end{tabular}
\label{compare}
\end{table}

% \begin{table}[t]
% \centering
% \caption{Comparison with other methods in different situations on MeshTalkSet\cite{richard2021meshtalk}. $E_{vl}$ and $E_{ve}$ represent the errors in lip and eye regions, respectively.The best results are highlighted (unit: millimeter). }
% \begin{tabular}{c|cc|cc}
% \hline
% \multirow{2}{*}{}&\multicolumn{2}{c|}{Noiseless audio}&\multicolumn{2}{c}{Noisy audio} \\ \cline{2-5} 
% &\multicolumn{1}{c|}{$E_{vl}$}&$E_{ve}$&\multicolumn{1}{c|}{$E_{vl}$}&$E_{ve}$\\ \hline
% VOCA\cite{cudeiro2019capture}&\multicolumn{1}{c|}{5.075}&3.224&\multicolumn{1}{c|}{5.273}&3.256\\ 
% GDPnet\cite{liu2021geometry}&\multicolumn{1}{c|}{-}&-&\multicolumn{1}{c|}{-}&-\\ 
% MeshTalk\cite{richard2021meshtalk}&\multicolumn{1}{c|}{6.385}&4.462&\multicolumn{1}{c|}{6.607}&4.669\\ 
% Ours&\multicolumn{1}{c|}{\textbf{4.676}}&\textbf{3.034}&\multicolumn{1}{c|}{\textbf{4.993}}&\textbf{3.040}\\ \hline
% \end{tabular}
% \label{compare_meshtalk}
% \end{table}

\begin{table}[t]
\centering
\caption{Comparison with other methods in different datasets on noisy audio. $E_{vl}$ and $E_{ve}$ represent the errors in lip and eye regions, respectively.The best results are highlighted (unit: millimeter). }
\begin{tabular}{c|cc|cc}
\hline
\multirow{2}{*}{}&\multicolumn{2}{c|}{VOCASET\cite{cudeiro2019capture}}&\multicolumn{2}{c}{MeshTalkSet\cite{richard2021meshtalk}} \\ \cline{2-5} 
&\multicolumn{1}{c|}{$E_{vl}$}&$E_{ve}$&\multicolumn{1}{c|}{$E_{vl}$}&$E_{ve}$\\ \hline
VOCA\cite{cudeiro2019capture}&\multicolumn{1}{c|}{6.659}&2.014&\multicolumn{1}{c|}{5.273}&3.256\\ 
GDPnet\cite{liu2021geometry}&\multicolumn{1}{c|}{6.480}&2.014&\multicolumn{1}{c|}{-}&-\\ 
MeshTalk\cite{richard2021meshtalk}&\multicolumn{1}{c|}{6.621}&2.104&\multicolumn{1}{c|}{6.607}&4.669\\ 
Ours&\multicolumn{1}{c|}{\textbf{6.051}}&\textbf{1.810}&\multicolumn{1}{c|}{\textbf{4.993}}&\textbf{3.040}\\ \hline
\end{tabular}
\label{compare_meshtalk}
\end{table}

\subsection{Audio-driven 3D facial animation}
In this section, we compare the results generated by our approach and the state-of-the-art methods, VOCA\cite{cudeiro2019capture}, GDPnet\cite{liu2021geometry} and MeshTalk\cite{richard2021meshtalk} in different aspects, including noiseless audio and noisy audio. As VOCA and GDPnet need specify the speaking style of subject, we choose speaking style randomly in this paper. 
In all experiments, we utilize the mean values of maximum mean square error of lip and eye regions as the evaluation metric, which is denoted by $E_{vl}$ and $E_{ve}$, respectively. All errors are measured on the test dataset. Furthermore, we also employ qualitative visual perception as a criterion, such as the range of mouth motion. Notably, to achieve a fair comparison, we only use the results generated at the first stage and do not predict posture variation in this part. 

% \begin{figure*}
% \centering
% \subfigure[Animation by audio1]{
% \begin{minipage}[t]{1\textwidth}
% \includegraphics[width = \textwidth]{data/compare/compare2.jpg}
% \end{minipage}
% }
% \vspace{-0.05in}
% \subfigure[Animation by audio2]{
% \begin{minipage}[t]{1\textwidth}
% \includegraphics[width = \textwidth]{data/compare/compare1.jpg}
% \end{minipage}
% }
% \caption{Normal audio evaluation experimental error map.}
% \label{numevaerrormap}
% \end{figure*}

%As mentioned above, we used the mean of the maximum mean square error of the lip and eye regions of the test set as the evaluation criterion. The best methods so far are VOCA, GDPnet and MeshTalk, but MeshTalk does not give training results and training codes on VOCASET, so we have to implement the training part by ourselves with our understanding of MeshTalk. In addition, since VOCA and GDPnet needs to manually input the person's identity during the testing process to represent the person's speaking style, if the manually selected style is far from the real style of the person in the test set, the results in terms of indicators will be poor. To try to avoid this effect, we fix a random seed and then sample random identities for each sentence in the test set.

\textbf{Evaluation on noiseless audios.}
In this paragraph, we compare our results with other methods on both VOCASET\cite{cudeiro2019capture} and MeshTalkSet\cite{richard2021meshtalk}. Note that, the results of GDPnet\cite{liu2021geometry} are not displayed in the MeshTalkSet as their released model is not suitable for the MeshTalkSet.
%\B{Recently, MeshTalk\cite{richard2021meshtalk} released their dataset, MeshTalkSet, so we compared our results with other methods on both VOCASET\cite{cudeiro2019capture} and MeshTalkSet. Unfortunately GDPnet\cite{liu2021geometry} does not provide a trainable code, so on MeshTalkSet we only compare with MeshTalk and VOCA.} 
Qualitative results are shown in Figs.~\ref{numevaerrormap1} and~\ref{numevaerrormap2}. In Fig.~\ref{numevaerrormap1}, we can find that VOCA, GDPnet and MeshTalk are not satisfactory in predicting facial subtle movements. For example, they all display large errors in the mouth regions. Figure \ref{numevaerrormap2} highlights the superiority of our method in predicting movement details. The accuracy of our method is significantly better than MeshTalk and VOCA in the mouth and eye regions. 
Table~\ref{compare} further illustrates the quantitative errors in the lip and eye regions on the above two dataset. Compared with these methods, the error of our approach is reduced by at least 0.7 
mm in terms of $E_{vl}$ in the VOCASET. For the error in the eye region, the error is reduced by nearly 0.2mm in the VOCASET.
In the MeshTalkSet, our method still shows a strong performance. 

\textbf{Evaluation on noisy audios.} In this paragraph, we test the effect of noise on different methods by adding Gaussian noise to the original audio.
%Noise is inevitable in our surrounding environment. Therefore, we test the robustness of these methods by adding Gaussian noise to the original audio and then comparing the sensitivity of each method to noise. 
The visualization results are shown in Figs.~\ref{noiseerrormap} and ~\ref{noiseerrormap2}. Table ~\ref{compare_meshtalk} shows the reconstruction errors on lip and eye regions. From the results, we can find that these methods are affected by noise to some extent. However, the proposed method still outperforms other methods under the interference of noise in terms of numerical errors. 
For example, compared with other methods, our error decreases by at least 0.44mm and 0.3mm in lip region on VOCASET and MeshTalkSet, respectively.%\B{and on the MeshTalkSet our mouth error is at least 0.3mm smaller}.
As the speech frames between Fig.~\ref{numevaerrormap1} and Fig.~\ref{noiseerrormap}, Fig.~\ref{numevaerrormap2} and Fig.~\ref{noiseerrormap2} are one-to-one in this paper, we can find that the noise mainly affects the lip, chin, eyes and cheek regions. 
%Thus, comparing Fig.~\ref{numevaerrormap1} and Fig.~\ref{noiseerrormap}, we can see that the main locations affected by noise are the lips, chin, and cheeks. Furthermore, the noise mainly affects the lip, chin and cheek regions.

%Table\ref{noise}, Figure\ref{noiseerrormap1} and Figure\ref{noiseerrormap2}, our method still outperforms other methods under the interference of noise, but it is more sensitive to outlier. This may be due to the greater influence of outliers on local features of speech. What's more, the speech frames are in a one-to-one correspondence in Figure\ref{numevaerrormap1}, Figure\ref{numevaerrormap2} and Figure\ref{noiseerrormap1}, Figure\ref{noiseerrormap2}, so we can see that the effect of noise on our method is mainly at the chin, which means that the movement of the lower lip is no longer as accurate.

\subsection{Pose evaluation}
%More recently related to our work on adding poses is the work of Zhang \etal~\cite{zhang20213d}, who also generated results with head poses. Unfortunately there is no published code for their work, so we cannot compare.

% At present, in the existing 3D face animation generation methods, no researchers propose to estimate head pose frame by frame during the animation generation process. In this section, we only give the results of our method. 

Figure~\ref{poseeva} visualizes some head motion results predicted at the second stage, which the first row is the results in VOCASET and the second is the results in MeshTalkSet. The proposed method can generate harmonious head posture based on given audio, as changes of posture usually happen during tone transitions. It indicates that the poses generated by our method from audio are reasonable, such as the 2\textit{th}, 4\textit{th} and 6\textit{th} synthetic results in the first row in Fig.~\ref{poseeva}. 

% \R{Note that since there is relatively little movement of the shoulder and neck during speaking, we used different factors for different positions of the template mesh when applying the poses.}

\begin{figure*}
\centering
\includegraphics[width=1\linewidth]{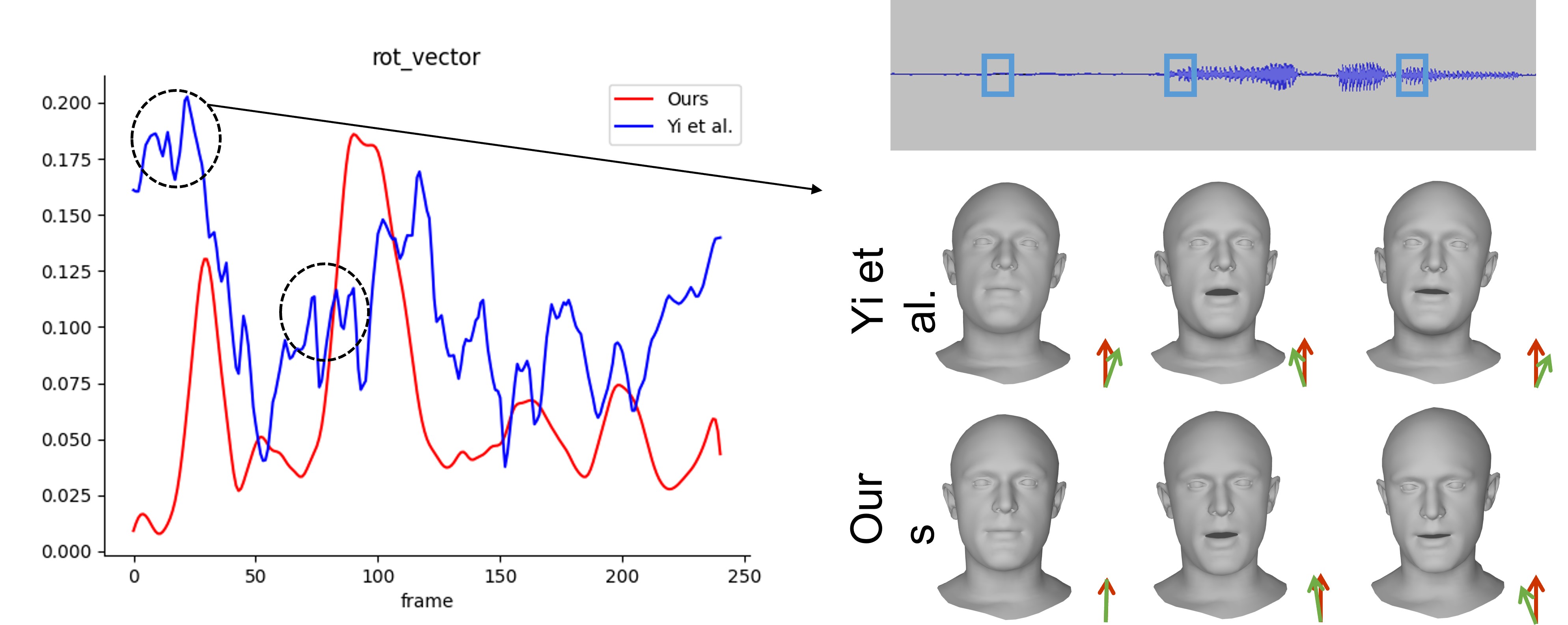}
\caption{Amplitude of pose variation between ours and Yi \etal~\cite{yi2020audio}. The horizontal axis represents the frames and the vertical axis represents the length of the head pose rotation vector.}
\label{compare_pose1}
\end{figure*}

% \begin{figure}
% \centering
% \includegraphics[width=1\linewidth]{data/compare/compare_pose2.jpg}
% \caption{Pose generation between the proposed method and Yi \etal~\cite{yi2020audio}.}
% \label{compare_pose2}
% \end{figure}

\begin{figure}
\centering
\includegraphics[width=1\linewidth]{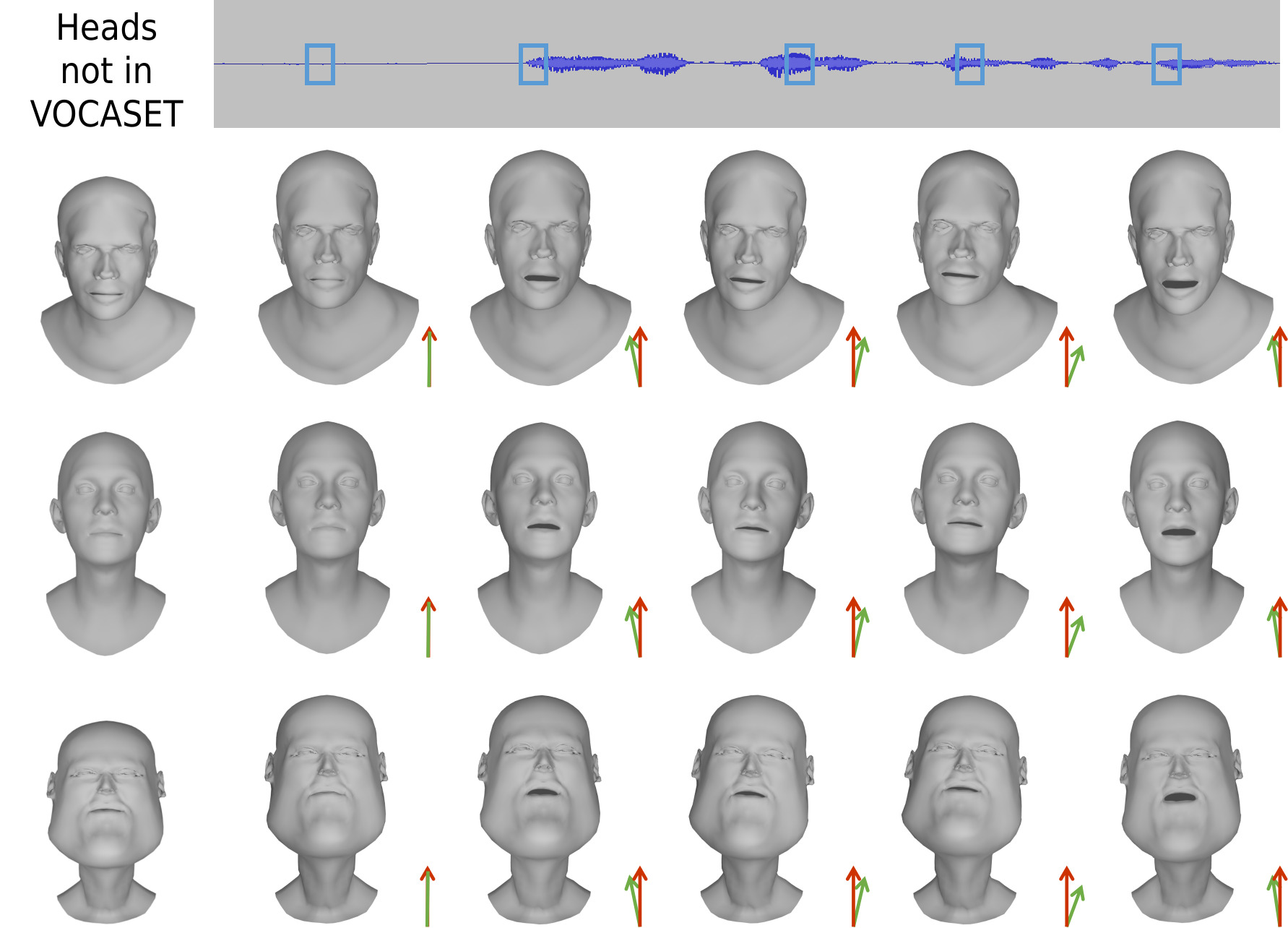}
\caption{Generalization on unseen head models which have the same topology structure with the meshes in VOCASET. The leftmost is the head meshes with significantly different appearance. Others are the speech-driven animation results.}
\label{othermodelE}
\end{figure}

% \begin{figure*}
% \centering
% \includegraphics[width=1\linewidth]{data/compare/compare_obj.jpg}
% \caption{Comparison in unseen subjects.}
% \label{otherobj}
% \end{figure*}

% \textbf{Comparison with other method.} 
Recently, Yi \etal~\cite{yi2020audio} proposed a parametric method to synthesis video about talking face. Their intermediate process could predict the variation of 3D head pose. Therefore, we test the performance of generating head pose between the proposed non-parametric method and the parametric one in this section. The comparison results are shown in Fig.~\ref{compare_pose1}. We can see that the variation of head pose in Yi \etal~\cite{yi2020audio} is frequent in a very short interval, see the marked circle from Fig.~\ref{compare_pose1} left. This phenomenon causes the head to shake continuously. As a contrast, our method produces a much smoother and more natural result of pose change. Figure~\ref{compare_pose1} right shows the synthetic results of head pose on a speech of about half one second. It can be seen that Yi \etal~\cite{yi2020audio} produces a violent, abnormal postural change, while our method produced a more consistent, downward motion to the left.

\textbf{User study.} We presented side-by-side clips of our approach versus Yi~\etal~\cite{yi2020audio} to a total of 53 participants and let them judge the task that which clip has more realistic head poses. For each individual, 39 pairs of short clips would been evaluated, each containing one sentence spoken by a subject from the VOCA test set. Participants could choose to favor one clip over the other.  The results of user study are shown in Tab.~\ref{user}, which indicate the competitiveness of our approach in generating postures. 71.17\% individuals surveyed believe that our method produces a more realistic head pose with better visual effects. The head posture changes produced by the Yi~\etal~\cite{yi2020audio} are too drastic and are disliked by many people. 

\begin{table}[t]
\centering
\caption{User study. Human participants were asked which of the two presented video clips presented a more realistic head pose while speaking.}
\begin{tabular}{ccc}
\hline
\multirow{2}{*}{}&\multicolumn{2}{c}{favorability}\\ \cline{2-3} 
&\multicolumn{1}{c}{competitor}&ours\\ \hline
Ours vs. Yi~\etal~\cite{yi2020audio}&28.83\%&71.17\% \\ \hline
\end{tabular}
\label{user}
\end{table}

\begin{figure*}
\centering
\subfigure[ French ]{
\begin{minipage}[t]{0.49\linewidth}
\includegraphics[width= \linewidth]{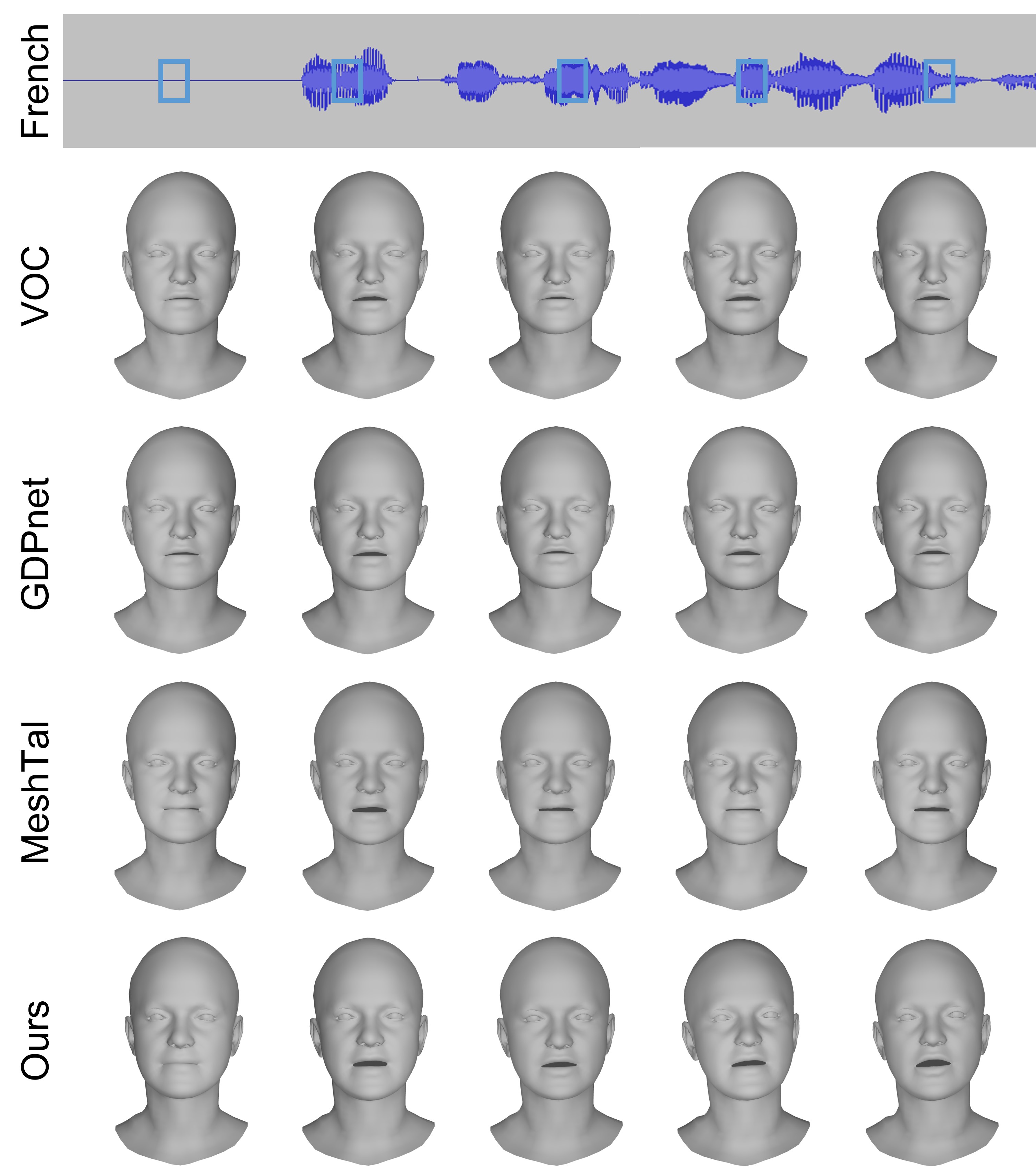}
\end{minipage}
}
\hspace{-.1in}
\subfigure[ German ]{
\begin{minipage}[t]{0.49\linewidth}
\includegraphics[width= \linewidth]{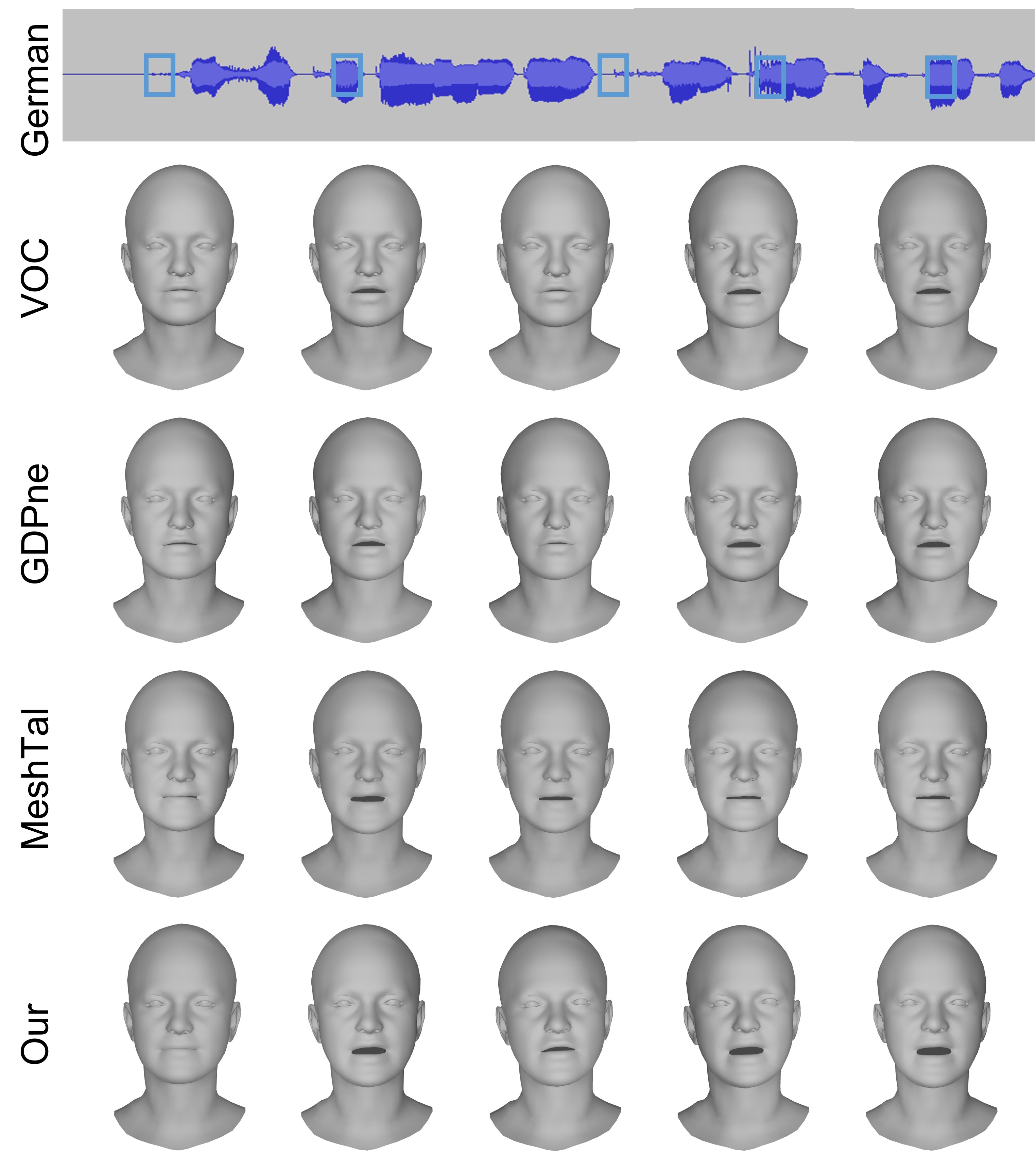}
\end{minipage}
}
\subfigure[ Japanese ]{
\begin{minipage}[t]{0.49\linewidth}
\includegraphics[width= \linewidth]{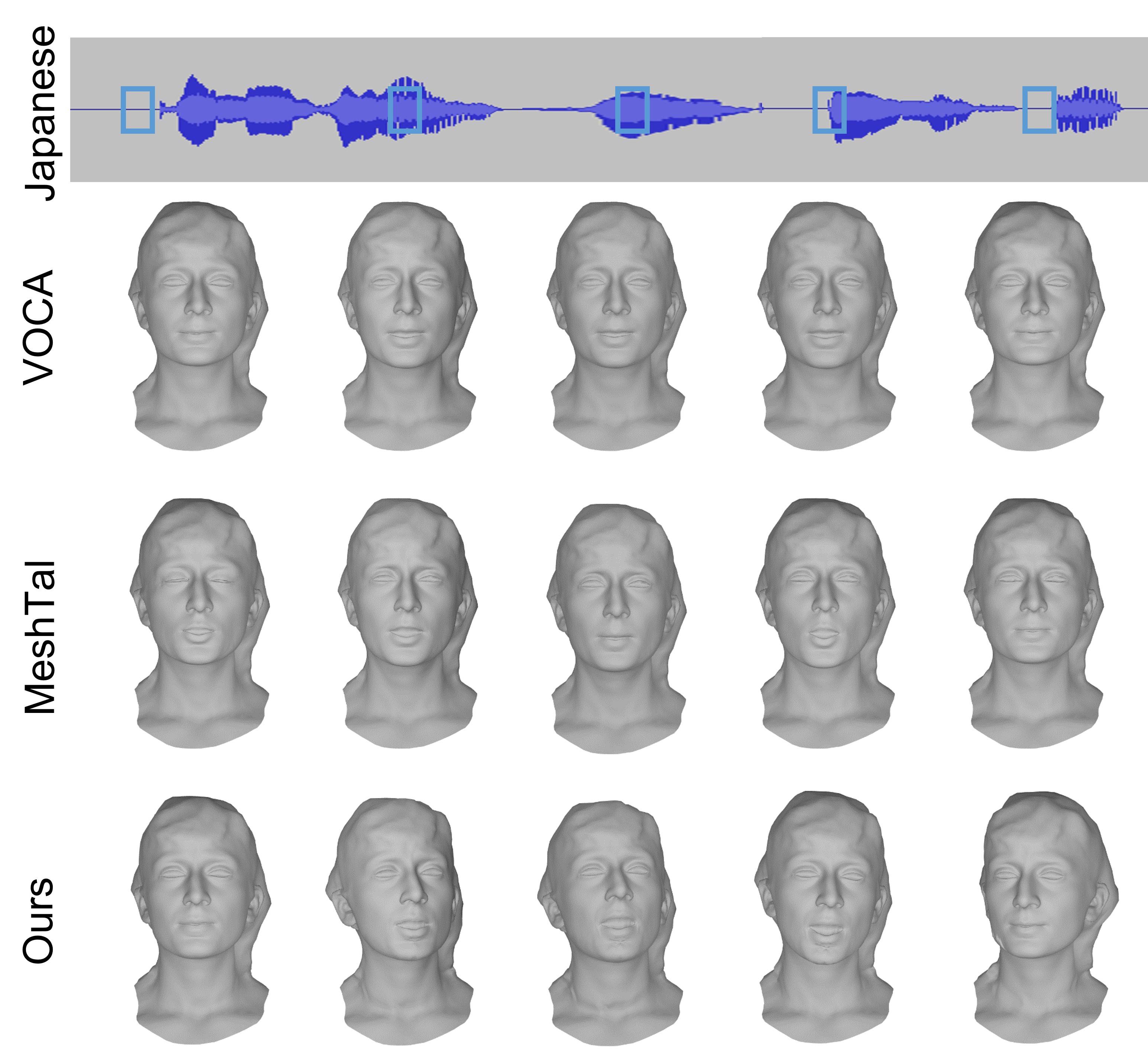}
\end{minipage}
}
\hspace{-.1in}
\subfigure[ Spanish ]{
\begin{minipage}[t]{0.49\linewidth}
\includegraphics[width= \linewidth]{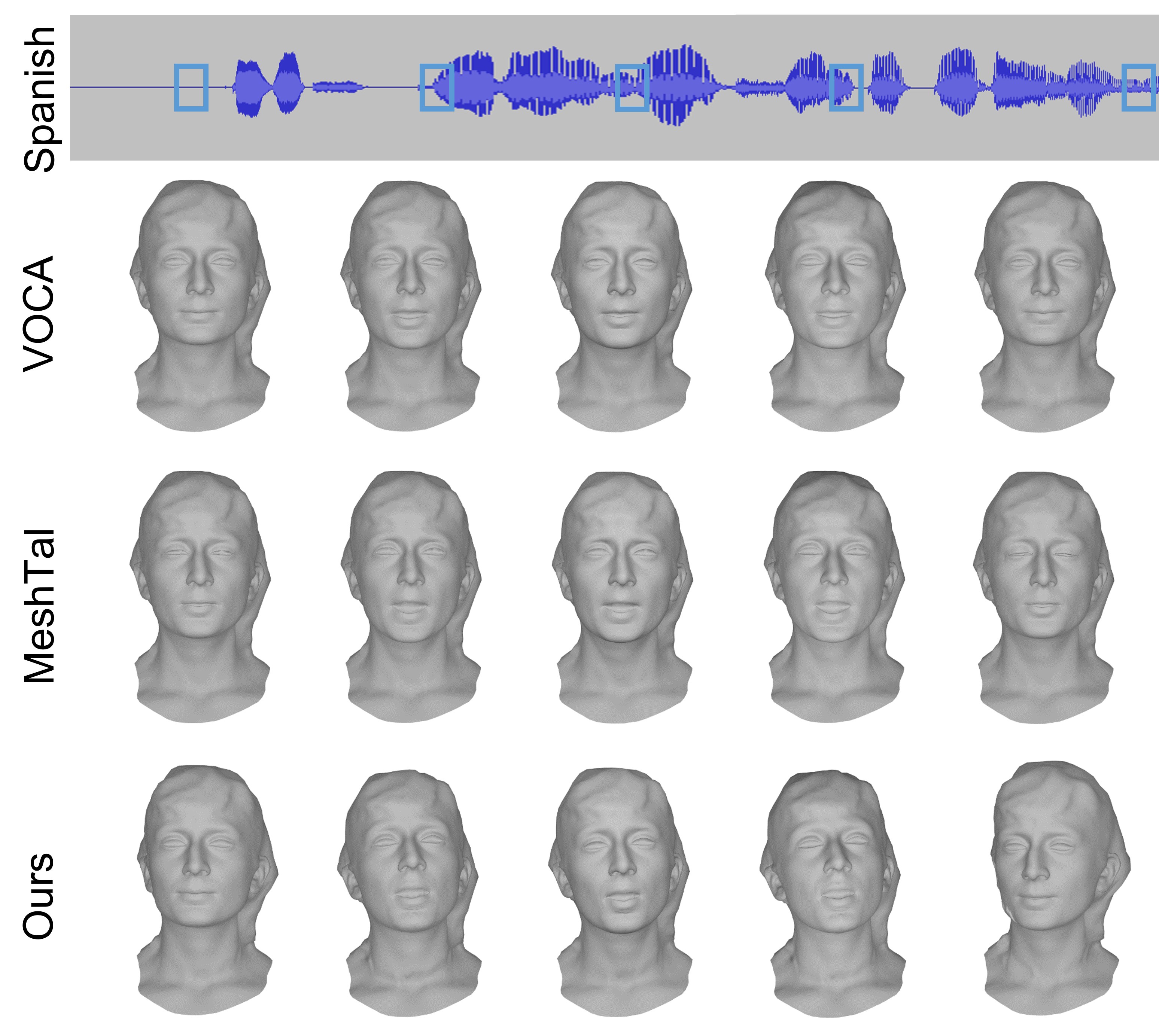}
\end{minipage}
}
\caption{Comparisons with other method among the cross-linguistic input.}
\label{compareotherlang}
\end{figure*}

\subsection{Generalization}
To evaluate the generalization of our method, we conduct some qualitative experiments in the following sections. 

\textbf{Generalization across unseen subjects.}
Our method can also perform well on unseen head meshes, which are significantly different from the training models in appearance. The experimental results are displayed in Fig.~\ref{othermodelE}. We can find that our method still has good performance on these head models. It further demonstrates the proposed method indeed learns the motion patterns of facial vertices with the help of hierarchical audio-vertex features, especially the speech-consistent mouth movements and head poses.
%The leftmost column shows some other FLAME\cite{li2017learning} head meshes, and then the right side shows the speech-driven animation process. Our method also produces high-quality results on non-standard datasets, especially with speech-consistent mouth movements and realistic head poses.

\begin{figure*}
\centering
\includegraphics[width=1\linewidth]{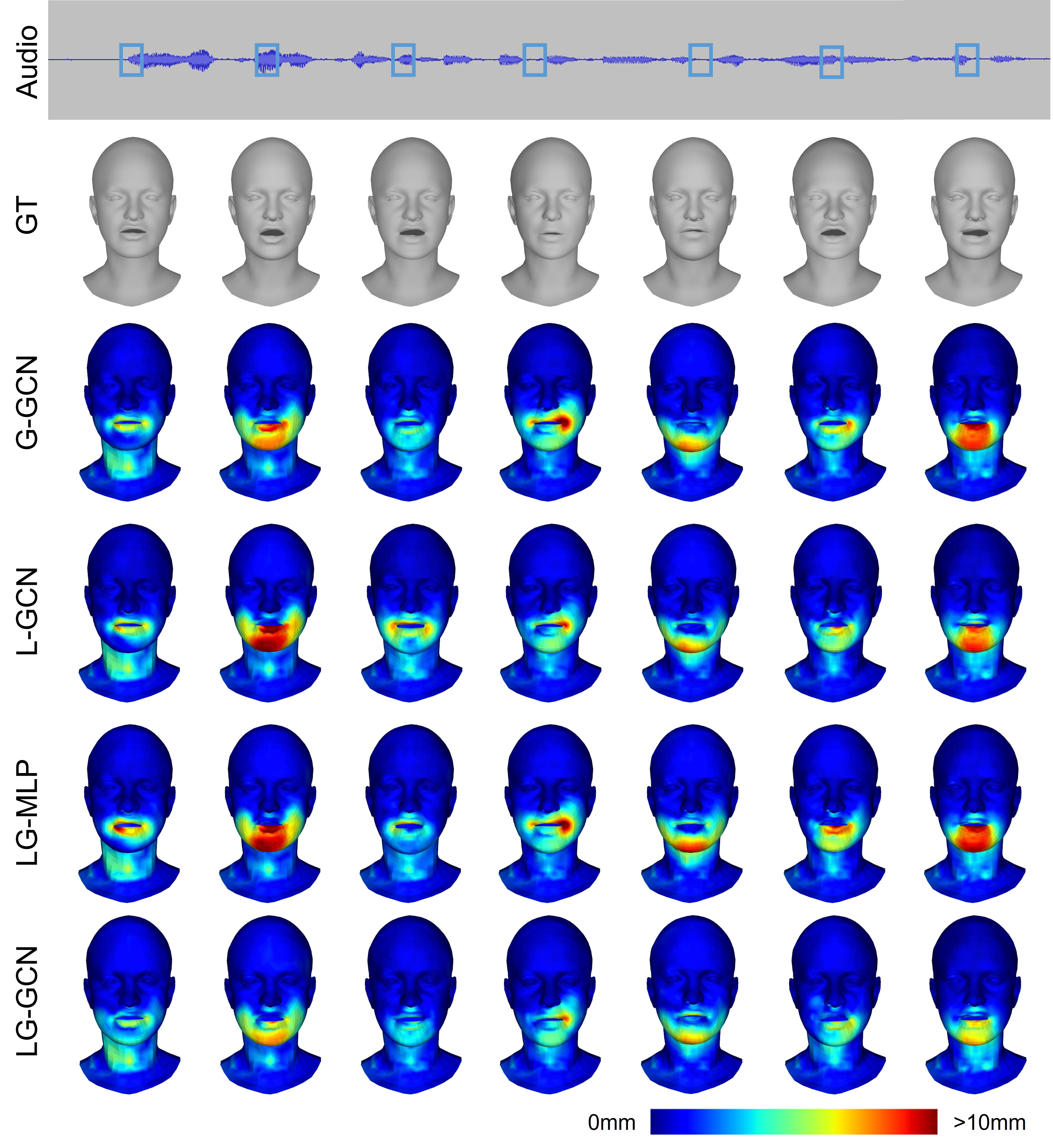}
\caption{Ablation experiments about extracting different features. The color maps give the distribution of vertex-to-vertex distance errors (unit: millimeter).}
\label{ablationerrormap}
\end{figure*}

\textbf{Generalization across language.}
The proposed method can also generate convincing animation results in other languages, such as French, German, Japanese and Spanish. We compare our method to other methods cross-linguistically and the results are shown in Fig.~\ref{compareotherlang}. Compared with other methods, we can find that our results have a tighter mouth closure in the frames where the audio signal keeps silent and a larger mouth opening in the places where the audio signal fluctuates a lot.

\subsection{Ablation experiments}
\subsubsection{Feature ablation experiments}

\begin{table}[t]
    \centering
    \caption{Ablation experimental results.}
    \begin{tabular}{c|c|c}
    \hline
    &$E_{vl}$ (mm)&$E_{ve}$ (mm) \\
    \hline
    L-GCN&5.300&\textbf{1.807}\\
    G-GCN&5.333&1.824\\
    LG-MLP&5.414&1.819\\
    LG-GCN&\textbf{5.261}&1.814\\
    \hline
    \end{tabular}
    \label{ablation}
\end{table}

We map speech signal into local and global features and concatenate them with mesh vertex index embedding as the input to the GCN network. For presentation convenience, we define this case as the Local+Global+GCN (LG-GCN) model, which is used to extract the hierarchical audio-vertex features. To test the effectiveness of the learned features and GCN, we designed several variants based on the proposed LG-GCN model,  the Local+GCN (L-GCN) model, the Global+GCN (G-GCN) model
%, the Local+MLP(L-MLP) model, the Global+MLP(G-MLP) model 
and the Local+Global+MLP(LG-MLP) model. We calculate the maximum value of the vertex-to-vertex mean squared error in the lip and eye regions per frame and use the mean value of the maximum values among all frames to evaluate the error for each variant.

Table~\ref{ablation} exhibits the metric results and some visualized results are shown in Fig.~\ref{ablationerrormap}. 
%At the same time, we also visualized the heat map Fig.~\ref{ablationerrormap} of the error results. 
On the one hand, compared with the G-GCN model, the L-GCN model works well in preserving facial details, such as the corners of the mouth in the fourth and sixth columns and the chin in the last column in Fig.~\ref{ablationerrormap}. This gives us an fact that audio local features are very useful in the generation of facial details. On the other hand, audio global features could constrain the distribution of all head vertices and may imply overall facial expression, such as the second and third columns. Therefore, we propose the hierarchical features fusion strategy by graph convolution network combining with the Fourier embedding features of the index of each vertex. Although using multilayer perceptron (MLP) could also fusion more features, their performance is not good, see the quantitative results in Tab.~\ref{ablation} and the qualitative results in the fifth row of Fig.~\ref{ablationerrormap}. As a contrast, GCN obtains satisfactory results because it makes full use of the prior information of mesh edges. Therefore, we finally choose LG-GCN as our module for extracting and fusing hierarchical audio-vertex features.

\subsubsection{Pose ablation experiments}

% \begin{figure}
% \centering
% \subfigure[The average of the absolute value of the change in pose in each direction between the previous and following frames.]{
% \begin{minipage}[t]{0.5\textwidth}
% \includegraphics[width = \textwidth]{data/pose/pose_mean.jpg}
% \end{minipage}
% }
% \subfigure[The average of the absolute values of the maximum values of postural change in each sentence.]{
% \begin{minipage}[t]{0.5\textwidth}
% \includegraphics[width = \textwidth]{data/pose/pose_max.jpg}
% \end{minipage}
% }
% \caption{pose ablation experiments result.}
% \label{poseablation}
% \end{figure}

\begin{figure}
\centering
\includegraphics[width=1\linewidth]{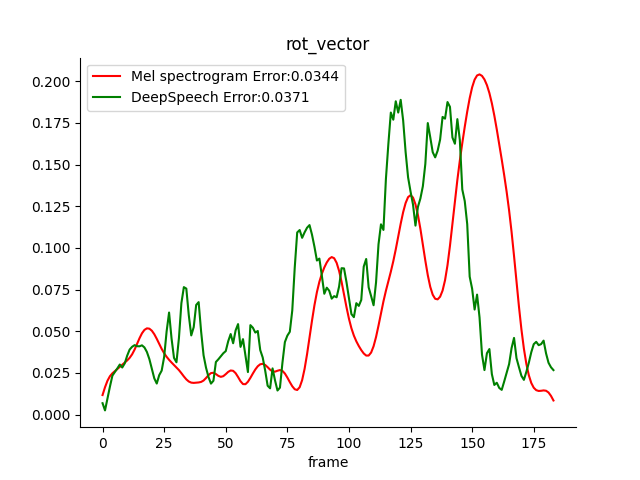}
\caption{Different head poses generated by Mel spectrogram and DeepSpeech features. The horizontal axis represents the frames and the vertical axis represents the length of the head pose rotation vector.}
\label{poseablation}
\end{figure}

% \begin{table}[t]
%     \centering
%     \caption{Modulus error of the rotation vector.}
%     \begin{tabular}{c|c}
%     \hline
%     &{Modulus error}\\
%     \hline
%     DeepSpeech&0.0371\\
%     Mel spectrogram&0.0344\\
%     \hline
%     \end{tabular}
%     \label{pose2real}
% \end{table}
% Following VOCA\cite{cudeiro2019capture}, GDPnet\cite{liu2021geometry} and MeshTalk\cite{richard2021meshtalk}, we use the same DeepSpeech feature $\mathbf{d_i}$ and Mel spectrogram feature $\mathbf{m_i}$. 

Following VOCA~\cite{cudeiro2019capture} and MeshTalk~\cite{richard2021meshtalk}, the input audio can be represented by DeepSpeech feature or Mel spectrogram feature. 
In this section, we conduct an ablation study to evaluate the performance of Mel spectrogram feature and DeepSpeech feature.
Figure~\ref{poseablation} illustrates the posture displacements of head pose predicted by these two features with a piece of voice.
We can find that using Mel spectrogram features can generate much smoother pose compared with DeepSpeech features, and the synthetic facial animation looks more coordinated. 
The experimental results validate that DeepSpeech features are closely related to the content (text), and affects detailed facial expression, while Mel spectrogram feature has strong relationship with head posture.
In addition, using Mel spectrogram feature to fit head pose also achieves higher accuracy, which can be shown in Fig.~\ref{poseablation}.

\section{Conclusion and Future Work}
We propose a novel pose-controllable audio-driven 3D facial animation synthesis method in this paper. A hierarchical audio-vertex feature is designed to predict the detailed facial expressions by integrating the global expression feature and local vertex-wise latent features. In order to conduct pose-controllable animation, we propose a novel pose attribute augmentation method based on 2D talking face synthesis. Numerical experiments demonstrate that the proposed method can produce more realistic facial animation with reasonable pose movements compared with the state-of-the-art methods.
One of the drawback of the proposed method is that we do not take into account the influence of emotions embedded in the audio. Therefore, in the future we expect to be able to develop realistic audio-driven emotional 3D face animation methods.

%-------------------------------------------------------------------------

{\small
\bibliographystyle{cvm}
\bibliography{cvmbib}
}

\end{document}